\documentclass[11pt,a4paper]{article} 
\usepackage[hyperref]{emnlp2018}
\usepackage{times}
\usepackage{latexsym}
\usepackage{paralist}
\usepackage{url}
\usepackage{todonotes}
\usepackage[normalem]{ulem}

\usepackage{graphicx}  
\usepackage{subcaption}

\aclfinalcopy 

\title{Do Explanations make VQA Models more Predictable to a Human?}

\author{\textbf{Arjun Chandrasekaran}\thanks{\,\,\,Denotes equal contribution.}$\,\,^{,1}$ \qquad 
\textbf{Viraj Prabhu}$^{*,1}$\qquad 
\textbf{Deshraj Yadav} $^{*,1}$ \qquad \\
\textbf{Prithvijit Chattopadhyay}$^{*,1}$ \qquad
\textbf{Devi Parikh}$^{1,2}$ \vspace{0.005\textwidth} \\
$^1${\fontsize{11}{12}\selectfont Georgia Institute of Technology}
\qquad $^2${\fontsize{11}{12}\selectfont Facebook AI Research}
\\
{\tt\small \{carjun, virajp, deshraj, prithvijit3, parikh\}@gatech.edu}}

\date{}

\begin{document}

\maketitle
\begin{abstract}
A rich line of research attempts to make deep neural networks more transparent by generating human-interpretable `explanations' of their decision process, especially for interactive tasks like Visual Question Answering (VQA).
In this work, we analyze if existing explanations indeed make a VQA model -- its responses as well as failures -- more predictable to a human. 
Surprisingly, we find that they do not. On the other hand, we find that human-in-the-loop approaches that treat the model as a black-box do.
\end{abstract}

\section{Introduction}

\noindent 
As technology progresses, we are increasingly collaborating with AI agents in \textit{interactive} scenarios where humans and AI work together as a team, e.g., in AI-assisted diagnosis, autonomous driving, etc. 
Thus far, AI research has typically only focused on the AI in such an interaction -- for it to be more accurate, be more human-like, understand our intentions, beliefs, contexts, and mental states. 

In this work, we argue that for human-AI interactions to be more effective, humans must also understand the AI's beliefs, knowledge, and quirks. 

Many recent works generate human-interpretable `explanations' regarding a model's decisions. These are usually evaluated offline based on whether human judges found them to be `good' or to improve trust in the model. However, their contribution in an interactive setting remains unclear. In this work, we evaluate the role of explanations towards making a model predictable to a human. 

\begin{figure}[t]
 \centering 
 \includegraphics[width=1\linewidth]{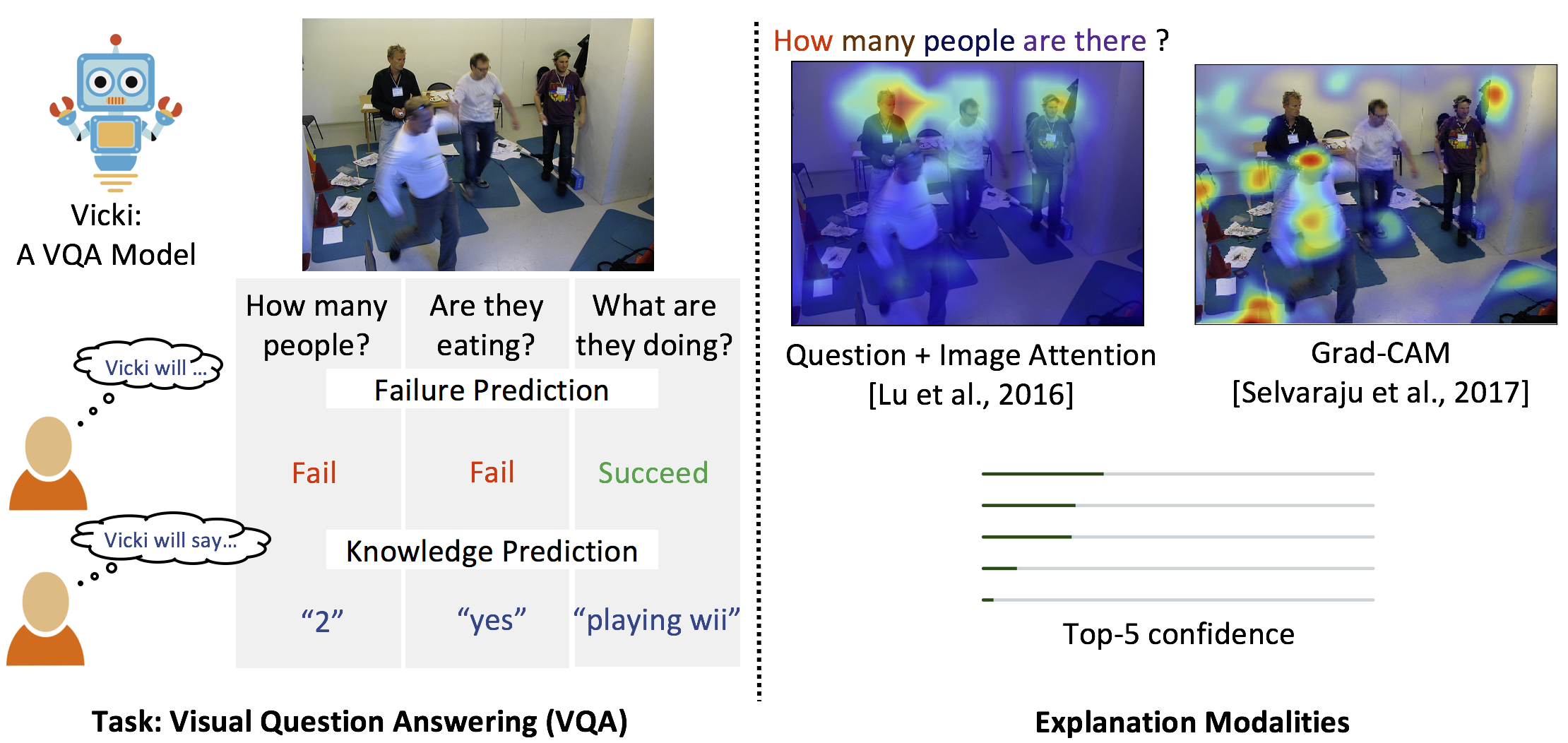}
 \vspace{-12pt}
 \caption{
    We evaluate the extent to which explanation modalities (right) and familiarization with a VQA model help humans predict its behavior -- its responses, successes, and failures (left).
 }
\vspace{-10pt}
\label{fig:teaser}
\end{figure}

We consider an AI trained to perform the multi-modal task of Visual Question Answering (VQA)~\cite{malinowski2014multi,antol2015vqa}, i.e., answering free-form natural language questions about images.
VQA is applicable to scenarios where humans actively elicit information from visual data, and naturally lends itself to human-AI interactions. 
We consider two tasks that demonstrate the degree to which a human understands their AI teammate (we call Vicki) -- Failure Prediction (FP) and Knowledge Prediction (KP). In FP, we ask subjects on Amazon Mechanical Turk to predict if Vicki will correctly answer a given question about an image. 
In KP, subjects predict Vicki's exact response. 

We aid humans in forming a mental model of Vicki by (1) familiarizing them with its behavior in a `training' phase and (2) exposing them to its internal states via various explanation modalities. We then measure their FP and KP performance. 

Our key findings are that (1) humans are indeed capable of predicting successes, failures, and outputs of the VQA model better than chance, (2) explicitly training humans to familiarize themselves with the model improves their performance, and (3) existing explanation modalities do not enhance human performance.

\section{Related Work}

\noindent
\textbf{Explanations in deep neural networks.} 
Several works generate explanations based on internal states of a decision process~\cite{zeiler2014visualizing,goyal2016towards}, while others generate justifications that are consistent with model outputs~\cite{ribeiro2016should, hendricks2016generating}.
Another popular form of providing explanations is to visualize regions in the input that contribute to a decision -- 
either by explicitly attending to relevant input regions~\cite{bahdanau2014neural, xu2015show}, or 
exposing implicit attention for predictions~\cite{selvaraju2016grad, zhou2016learning}. 

\noindent
\textbf{Evaluating explanations.} 
Several works evaluate the role of explanations in developing trust with users~\cite{cosley2003seeing,ribeiro2016should} or helping them achieve an end goal~\cite{narayanan2018humans,kulesza2012tell}. 
Our work, however, investigates the role of machine-generated explanations in improving the \textit{predictability} of a VQA model. 

\noindent 
\textbf{Failure prediction.} While~\citet{bansal2014towards} and~\citet{zhang2014predicting} predict failures of a model using simpler statistical models, we explicitly train a person to do this. 

\noindent
\textbf{Legibility.}~\citet{dragan2013legibility} describe the intent-expressiveness of a robot as its trajectory being expressive of its goal. Analogously, we evaluate if explanations of the intermediate states of a VQA model are expressive of its output.

\noindent
\textbf{Humans adapting to technology.} ~\citet{wang2016learning} and ~\citet{pelikan2016nao} observe humans' strategies while adapting to the limited capabilities of an AI in interactive language games.
In our work we explicitly measure to what extent humans can form an accurate model of an AI, and the role of familiarization and explanations.

\section{Setup}
\label{sec:setup}

\textbf{Agent.}
We use the VQA model by~\citeauthor{lu2016hierarchical}~\shortcite{lu2016hierarchical} as our AI agent (that we call Vicki). The model processes the question at multiple levels of granularity (words, phrases, entire question) and at each level, has explicit attention mechanisms on both the image and the question\footnote{We use question-level attention maps in our experiments.}. 
It is trained on the train split of the VQA-1.0 dataset~\cite{antol2015vqa}. Given an image and a question about the image, it outputs a probability distribution over 1000 answers. 
Importantly, the model's image and question attention maps provide access to its `internal states' while making a prediction. 

Vicky is \textit{quirky} at times, i.e., has biases, albeit in a predictable way.~\citet{agrawal2016analyzing} outlines several such quirks.
For instance, Vicki has a limited capability to understand the image
-- when asked the color of a small object in the scene, say a soda can, it may simply respond with the most dominant color in the scene. Indeed, it may answer similarly even if no soda can is present, i.e. if the question is irrelevant. 

Further, Vicki has a limited capability to understand free-form natural language, and in many cases, answers questions based only on the first few words of the question. 
It is also generally poor at answering questions requiring ``common sense'' reasoning.
Moreover, being a discriminative model, Vicki has a limited vocabulary (1k) of answers.
Additionally, the VQA 1.0 dataset contains label biases; therefore, the model is very likely to answer ``white' to a ``what color'' question~\cite{goyal2016making}. 

\begin{figure}[t]
 \centering 
 \includegraphics[width=1\linewidth]{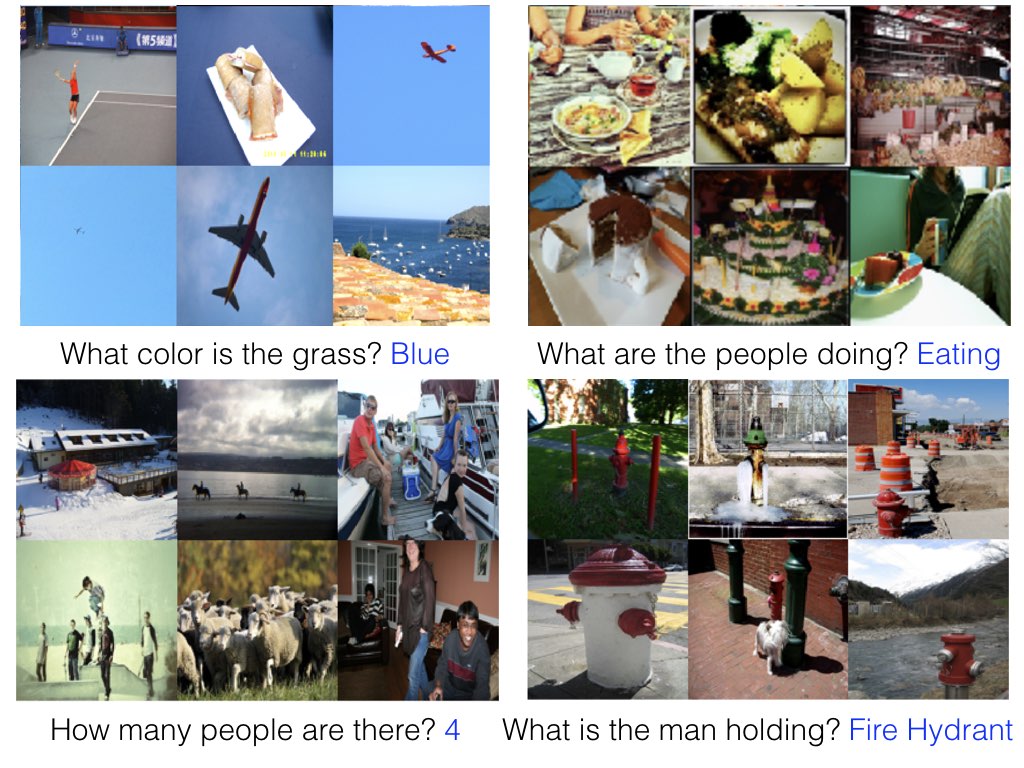}
 \vspace{-20pt}
 \caption{These montages highlight some of Vicki's quirks. For a given question, Vicki has the same response to each image in a montage. Common visual patterns (that Vicki presumably picks up on) within each montage are evident.}
 \vspace{-15pt}
\label{fig:montages}
\end{figure}

To get a sense for this, see Fig.~\ref{fig:montages} which depicts a clear pattern. In top-left, even when there is no grass, Vicki tends to latch on to one of the dominant colors in the image. For top-right, even when there are no people in the image, it seems to respond with what people could \emph{plausibly} do in the scene if they were present.
In this work, we measure to what extent lay people can pick up on these quirks by interacting with the agent, and whether existing explanation modalities help do so.

\noindent
\textbf{Tasks:} 
\label{subsec:tasks}
\textbf{Failure Prediction (FP).} 
Given an image and a question about the image, we measure how well a person can predict if Vicki will successfully answer the question.
A person can presumably predict the failure modes of Vicki well if they have a good sense of its strengths and weaknesses. 

\begin{figure*}[!t]
\setlength{\fboxsep}{0pt}
\setlength{\fboxrule}{0pt}
	\begin{subfigure}[t]{3in}	\fbox{\includegraphics[width=3in]{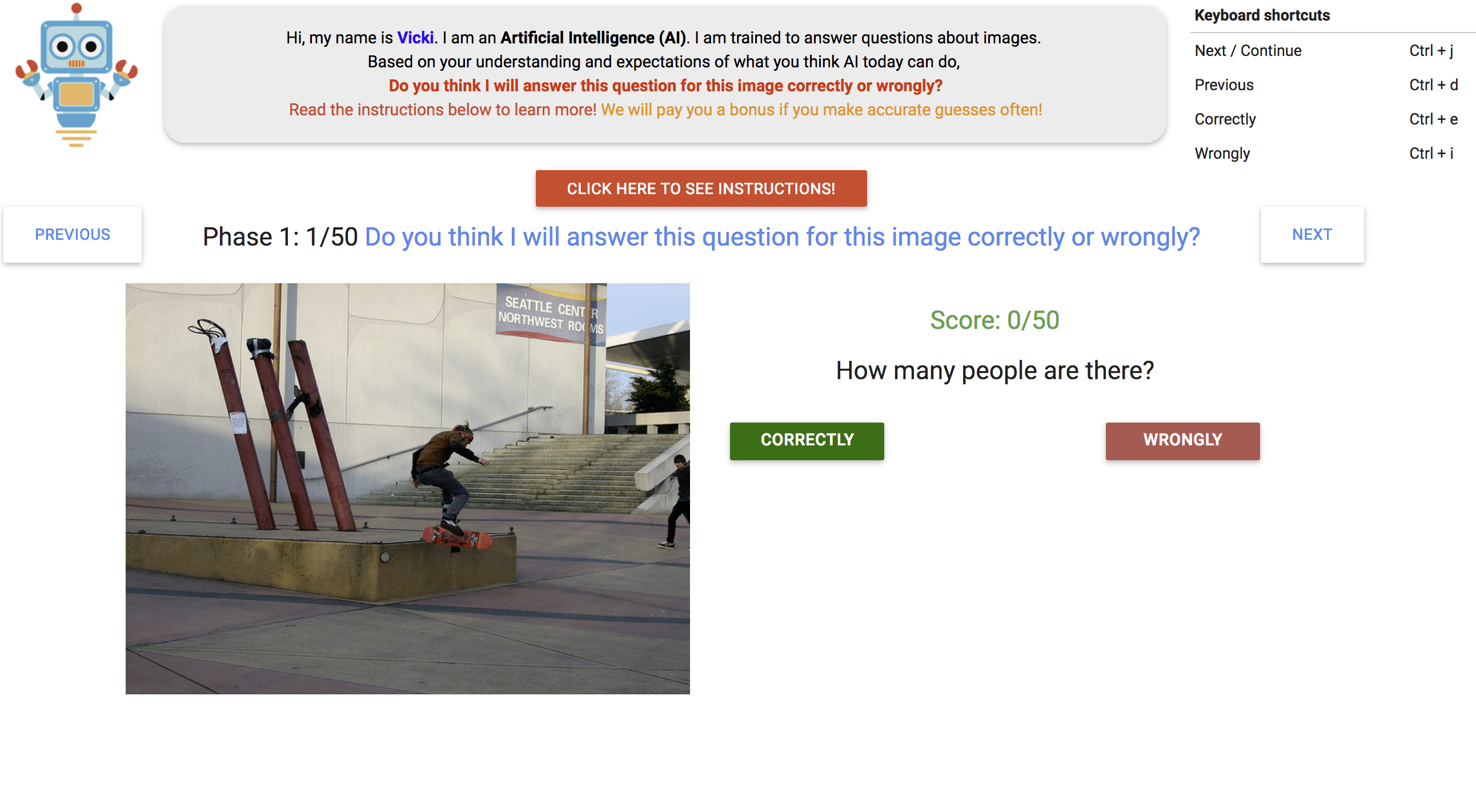}}
    \vspace{-12pt}
		\caption{The Failure Prediction (FP) interface.}
        \label{fig:fp_interface}
	\end{subfigure}
	\begin{subfigure}[t]{3in}
		\fbox{\includegraphics[width=3in]{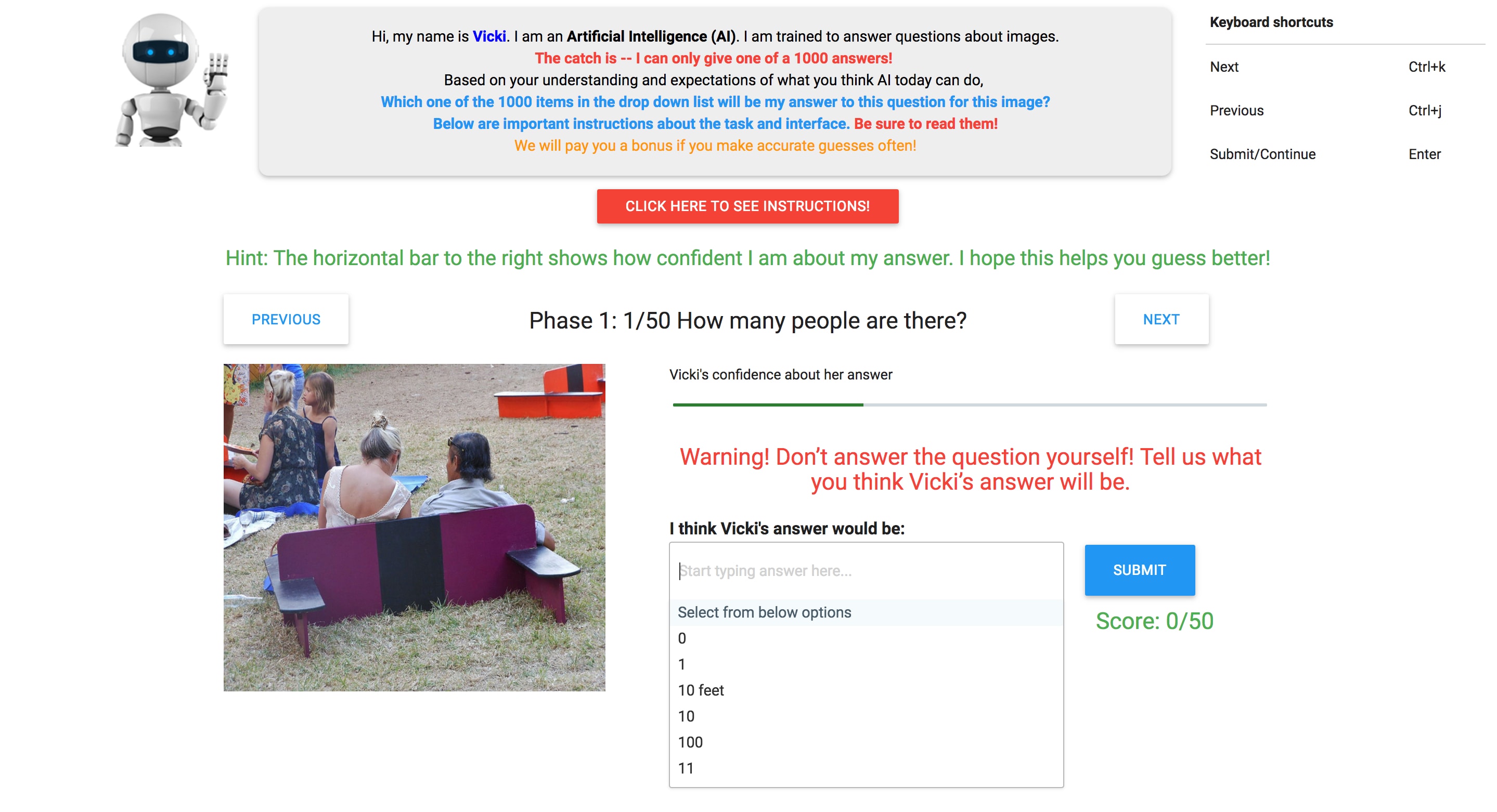}}
\vspace{-12pt}
		\caption{The Knowledge Prediction (KP) interface.}
        \label{fig:kp_interface}
	\end{subfigure}
    \vspace{-5pt}
\caption{(a) A person guesses if a VQA model (Vicki) will answer this question for this image correctly or wrongly. (b) A person guesses what Vicki's exact answer will be for this QI--pair.} 
\label{fig:interfaces}
\vspace{-6pt}
\end{figure*}

\noindent
\textbf{Knowledge Prediction (KP).} In this task, we aim to obtain a fine-grained measure of a person's understanding of Vicki's behavior. Given a QI--pair, a subject guesses Vicki's exact response from a set of its output labels. 
Snapshots of our interfaces can be seen in Fig.~\ref{fig:interfaces}.
\section{Experimental Setup}

In this section we investigate ways to make Vicki's behavior more predictable to a subject. 
We approach this by -- providing instant feedback about Vicki's actual behavior on each QI pair once the subject responds, and exposing subjects to various explanation modalities that reveal Vicki's internal states before they respond. 

\noindent
\textbf{Data.} We identify a subset of questions in the VQA-1.0~\cite{antol2015vqa} validation split that occur more than 100 times. We select 7 diverse questions\footnote{What kind of animal is this? What time is it? What are the people doing? Is it raining? What room is this? How many people are there? What color is the umbrella?} from this subset that are representative of the different types of questions (counting, yes/no, color, scene layout, activity, etc.) in the dataset.  For each of the 7 questions, we sample a set of 100 images.
For FP, the 100 images are random samples from the set of images on which the question was asked in VQA-1.0 val. 
For the KP task, these 100 images are random images from VQA-1.0 val. ~\citet{ray2016question} found that randomly pairing an image with a question in the VQA-1.0 dataset results in about 79\% of pairs being irrelevant. This combination of relevant and irrelevant QI-pairs allows us to test subjects' ability to develop a robust understanding of Vicki's behavior across a wide variety of inputs.

\noindent 
\textbf{Study setup.} We conduct our studies on Amazon Mechanical Turk. Each task (HIT) comprises of 100 QI-pairs where for simplicity (for the subject), a single question is asked across all 100 images. The annotation task is broken down into a train and test phase of 50 QI-pairs each. Over all settings, 280 workers took part in our study (1 unique worker per HIT), resulting in 28k human responses. Subjects were paid an average of $\$3$ base plus $\$0.44$ performance bonus, per HIT.

\noindent
There are challenges involved in scaling data-collection in this setting: (1) Due to the presence of separate train and test phases, our AMT tasks tend to be unusually long (mean HIT durations across the tasks of FP and KP $=10.11\pm1.09$ and $24.49\pm1.85$ min., respectively).  Crucially, this also reduces the subject pool to only those willing to participate in long tasks. (2) Once a subject participates in a task, they cannot do another because their familiarity with Vicki would leak over. This constraint causes our analyses to require as many subjects as tasks. Since work division in crowdsourcing tasks follows a Pareto principle~\cite{little2009many}, this makes data collection very slow. 
In light of these challenges, we focus on a small set of questions to systematically evaluate the role of training and exposure to Vicki's internal states.

\vspace{-2pt}
\subsection{Evaluating the role of familiarization} 
\label{subsec:feedback} 
\vspace{-1pt}

To familiarize subjects with Vicki, we provide them with instant feedback during the train phase. Immediately after a subject responds to a QI--pair, we show them whether Vicki actually answered the question correctly or not (in FP) or what Vicki's response was (in KP), along with a running score of how well they are doing. 
Once training is complete, no further feedback is provided and subjects are asked to make predictions for the test phase. At the end, they are shown their score and paid a bonus proportional to the score.

\begin{figure*}[t]
 \centering 
 \includegraphics[width=1\linewidth]{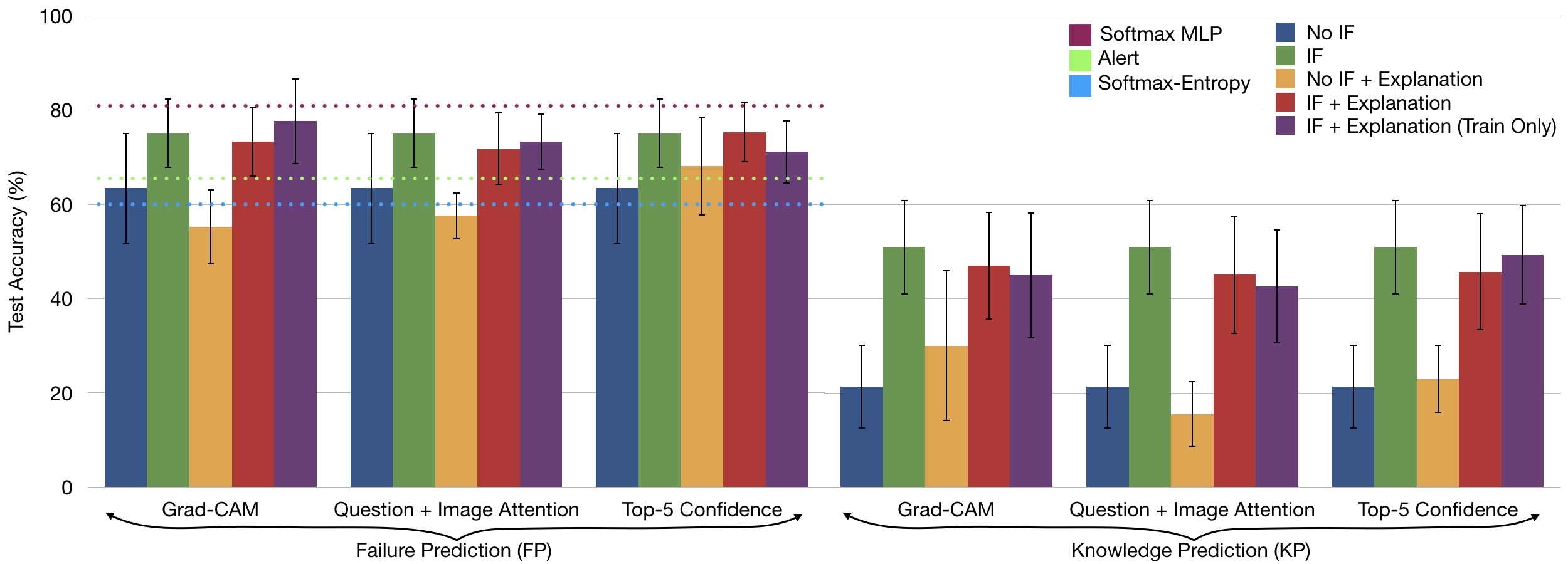}
 \vspace{-20pt}
 \caption{Average performance across subjects for Failure Prediction and Knowledge Prediction, across different settings: with or without (1) Instant feedback (IF) in the train phase, and (2) an explanation modality. 
 Explanation modalities are shown in both train and test phases unless stated otherwise.
 Error bars are 95\% confidence intervals from 1000 bootstrap samples. Note that the dotted lines are various machine approaches applied to FP. 
 }
\label{fig:xai}
\end{figure*}

\noindent
\textbf{Failure Prediction.} In FP, always guessing that Vicki answers `correctly' results in 58.29\% accuracy, while subjects do slightly better and achieve 62.66\% accuracy, even without prior familiarity with Vicki (No Instant Feedback (IF)). 
Further, we find that subjects that receive training via instant feedback (IF) achieve 13.09\% higher mean accuracies than those who do not (see Fig~\ref{fig:xai}; IF vs No IF for FP (left)). 

\noindent
\textbf{Knowledge Prediction.} In KP, answering each question with Vicki's most popular answer overall (`no') would lead to an accuracy of 13.4\%. Additionally, answering each question with its most popular answer \emph{for that question}
leads to an accuracy of 31.43\%. Interestingly, subjects who are unfamiliar with Vicki (No IF) achieve 21.27\% accuracy -- better than the most popular answer overall, but worse than the question-specific prior over its answers. The latter is understandable as subjects unfamiliar with Vicki do not know which of its 1000 possible answers the model is most likely to predict for each question. 

We find that mean performance in KP with IF is 51.11\%, 29.84\% higher than KP without IF (see Fig~\ref{fig:xai}; IF vs No IF for KP (right)). 
It is apparent that just from a few (50) training examples, subjects succeed in building a mental model of Vicki's behavior that generalizes to new images. 
Additionally, the 29.84\% improvement over No IF for KP is significantly larger than that for FP (13.09\%). This is understandable because a priori (No IF), KP is a much harder task as compared to FP due to the increased space of possible subject responses given a QI-pair, and the combination of relevant and irrelevant QI-pairs in the test phase. 

Questions such as `Is it raining?' have strong language priors -- to these Vicki often defaults to the most popular answer (`no'), irrespective of image. On such questions, subjects perform considerably better in KP once they develop a sense for Vicki's inherent biases via instant feedback. For open-ended questions like `What time is it?', feedback helps subjects (1) narrow down the 1000 potential options to the subset that Vicki typically answers with  -- in this case time periods such as `daytime' rather than actual clock times and (2) identify correlations between visual patterns and Vicki's answer. In other cases like `How many people are in the image?' the space of possible answers is clear a priori, but after IF subjects realize that Vicki is bad at detailed counting and bases its predictions on coarse signals of the scene layout.
\vspace{-5pt}
\subsection{Evaluating the role of explanations} 
\label{subsec:x_modalities}
\vspace{-1pt}

In this setting, we show subjects an image, a question, and one of the explanation modalities described below.
We experiment with 3 qualitatively different modalities (see Fig.~\ref{fig:teaser}, right):
\noindent
\textbf{Confidence of top-5 predictions.} 
We show subjects Vicki's confidence in its top-5 answer predictions from its vocabulary as a bar plot (of course, we do not show the actual top-5 predictions).
\textbf{Attention maps.} 
Along with the image we show subjects the spatial attention map over the image and words of the question which indicate the regions that Vicki is looking at and listening to, respectively.  
\textbf{Grad-CAM.} 
We use the CNN visualization technique by~\citeauthor{selvaraju2016grad}~\shortcite{selvaraju2016grad}, using the (implicit) attention maps corresponding to Vicki's most confident answer.

\noindent
\textbf{Automatic approaches}.
We also evaluate automatic approaches to detect Vicki's failure from its internal states. We find that both, a decision stump on Vicki's confidence in its top answer, and on the entropy of its softmax output, result in an FP accuracy of 60\% on our test set. 
A Multi-layer Perceptron (MLP) trained on Vicki's output 1000-way softmax to predict success vs failure, achieves an FP accuracy of 81\%. 
Training on just top-5 softmax outputs achieves an FP accuracy of 61.43\%.
An MLP which takes as input question features (average word2vec embeddings~\cite{mikolov2013distributed} of words in the question) concatenated with image features (fc7 from VGG-19) to predict success vs failure (which we call ALERT following~\cite{zhang2014predicting}) achieves an FP accuracy of 65\%. Training an MLP on identical question features as above but concatenated with Grad-CAM saliency maps leads to FP accuracy of 73.14\%\footnote{These methods are trained on 66\% of VQA-1.0 val. The remaining data is used for validation.}. 
Note that we report machine results only to put human performance in perspective. We do not draw inferences about the relative capabilities of both. 

\noindent
\textbf{Results.} Average performance of subjects in the test phases of FP and KP, for different experimental settings are summarized in Fig.~\ref{fig:xai}. 
In the first setting, we show subjects an explanation modality with instant feedback (IF+Explanation).
For reference, also see performance of subjects provided with IF and no explanation modality (IF). 

We observe that on both FP and KP, subjects who received an explanation along with IF show no statistically significant difference in performance compared to those who did not. We see in Fig.~\ref{fig:xai}, that both bootstrap based standard error (95\% confidence intervals) overlap significantly. 

Seeing that explanations in addition to IF does not outperform an IF baseline, we next measure whether explanations help a user not already familiar with Vicki via IF. That is, we evaluate if explanations help against a No IF baseline by providing an explanation only in the \textit{test} phase, and no IF (see Fig.~\ref{fig:xai}; No IF + Explanation). 
Additionally, we also experiment with providing IF and an explanation \textit{only} during the train phase (see Fig.~\ref{fig:xai}; IF + Explanation (Train Only)), to measure whether access to internal states during training can help subjects build better intuitions for model behavior without needing access to internal states at test time. 
In both settings however, we observe no statistically significant difference in performance over the No IF and IF baselines, respectively.\footnote{When piloting the tasks ourselves, we found it easy to `overfit' to the explanations and hallucinate patterns.}
\section{Conclusion}

As technology progresses, human-AI teams are inevitable. We argue that for these teams to be more effective, we should also be pursuing research directions to help humans understand the strengths, weaknesses, quirks, and tendencies of AI.  We instantiate these ideas in the domain of Visual Question Answering (VQA), by proposing two tasks that help measure how well a human `understands' a VQA model (we call Vicki) -- Failure Prediction (FP) and Knowledge Prediction (KP). 
We find that lay people indeed get better at predicting Vicki's behavior using just a few `training' examples, but surprisingly, existing popular explanation modalities do not help make its failures or responses more predictable. 
While previous works have typically assessed their interpretability or their role in improving human trust, our preliminary hypothesis is that these modalities may not yet help performance of human-AI teams in a goal-driven setting.
Clearly, much work remains to be done in developing improved explanation modalities that can improve human-AI teams.

Future work involves closing the loop and evaluating the extent to which improved human performance at FP and KP translates to improved success of human-AI teams at accomplishing a shared goal. Co-operative human-AI games may be a natural fit for such an evaluation.

\noindent
\textbf{Acknowledgements.} We thank Satwik Kottur for his help with data analysis, and for many fruitful discussions. We thank Aishwarya Agrawal and Akrit Mohapatra for their help with experiments. We would also like to acknowledge the workers on Amazon Mechanical Turk for their effort. 
This work was supported in part by NSF, AFRL, DARPA, Siemens, Google, Amazon, ONR YIP and ONR Grants N00014-16-1-2713.

\newpage

\bibliographystyle{acl_natbib_nourl}
\bibliography{emnlp2018}

\appendix

\renewcommand{\thesection}{\Roman{section}}
\renewcommand{\thesubsection}{\thesection.\Roman{subsection}}

\section*{Appendix}
The appendix is organized as follows. 
We first present further details regarding the role of familiarization which was presented in~\ref{subsec:feedback}. 
Following this, in Sec.~\ref{sec:visual_recognition_scenarios}, we discuss various visual recognition scenarios in which a human might rely on an AI, and motivate the need for building a model of the AI in such scenarios. 
Next, we discuss the Failure Prediction (FP) and Knowledge Prediction (KP) tasks (described earlier in brief in Sec.~\ref{subsec:tasks}). 
We then provide qualitative examples of montages that highlight the quirks (see Agent, Sec.~\ref{sec:setup}) which make Vicki predictable, and additionally share insights on Vicki from subjects who completed the tasks, in Sec.~\ref{sec:vickis_quirks}. 
Finally, we describe an AMT survey we conducted to gauge public perception of AI, and provide a list of questions and qualitative analyses of results.

\section{Analysis of the Role of Familiarization}

Just as an anecdotal point of reference, we also conducted experiments across experts with varying degrees of familiarity with agents like Vicki. We observed that a VQA researcher had an accuracy of 80\% versus a computer vision (but not VQA) researcher who had 60\% in a shorter version of the FP task without instant feedback. Clearly, familiarity with Vicki appears to play a critical role in how well a human can predict its oncoming failures or successes. 

\section{Visual Recognition Scenarios}
\label{sec:visual_recognition_scenarios}
In general, one might wonder why a human would need Vicki to answer questions if they are already looking at the image. This may be true for the VQA dataset, but outside of that there are scenarios where the human either does not know the answer to a question of interest (e.g., the species of a bird), or the amount of visual data is so large (e.g., long surveillance videos) that it would be prohibitively cumbersome for them to sift through it. Note that even in this scenario where the human does not know the answer to the question, a human who understands Vicki's failure modes from past experience would know when to trust its decision. For instance, if the bird is occluded, or the scene is cluttered, or the lighting is bad, or the bird pose is odd, Vicki will likely fail. Moreover, the idea of humans predicting the AI's failure also applies to other scenarios where the human may not be looking at the image, and hence needs to work with Vicki (e.g., blind user, or a human working with a tele-operated robot). In these cases too, it would be useful for the human to have a sense for the contexts and environments and/or kinds of questions for which Vicki can be trusted. In this work, as a first step, we focus on the first scenario where the human is looking at the image and a question while predicting Vicki's failures and responses.

\section{Tasks and Interfaces}

Our proposed tasks of FP and KP are designed to measure a human's understanding of the capabilities of an AI agent such as Vicki. As mentioned before, the tasks are especially relevant to human-AI teams since they are analogous to measuring if a human teammate's trust in an AI teammate is well-calibrated, and if the human can estimate the behavior of an AI in a specific scenario.

\noindent
\textbf{Failure Prediction. }Recall that in FP, given an image and a question about the image, we measure how accurately a person can predict if Vicki will successfully answer the question. A collaborator who performs well on this task can accurately determine whether they should trust Vicki's response to a question about an image. Please see a snapshot of the FP interface in Fig.~\ref{fig:fp_interface}. Note that we do not show the human what Vicki's predicted answer is. 

\noindent
\textbf{Knowledge Prediction. }In KP, given an image and a question, a person guesses Vicki's exact response (answer) from a set of its output labels (vocabulary). Recall that Vicki can only say one of a 1000 things in response to a question about an image. Please see a snapshot of the KP interface in Fig.~\ref{fig:kp_interface}. We provide subjects a convenient dropdown interface with autocomplete to choose an answer from Vicki's vocabulary of 1000 answers.

In FP, a good understanding of Vicki's strengths and weaknesses might lead to good human performance. However, KP requires a deeper understanding of Vicki's behavior, rooted in its quirks and beliefs. In addition to reasoning about Vicki's failure modes, one has to guess its exact response for a given question about an image. Note that KP measures subjects' ability to take reality (the image the subject sees) and translate it to what Vicki might say. High performance at KP is likely to correlate to high performance at the reverse task -- take what Vicki says and translate it to what the image really contains. This can be very helpful when the visual content (image) is not directly available to the user. Explicitly measuring this is part of future work. A person who performs well at KP has likely successfully modeled a more fine-grained behavior of Vicki than just modes of success or failure. In contrast to typical efforts where the goal is for AI to approximate human abilities, KP involves measuring a human’s ability to approximate a neural network’s behavior!

We used different variants of the base interfaces (see Fig.~\ref{fig:fp_interface} and~\ref{fig:kp_interface}) for both Failure Prediction and Knowledge Prediction tasks on Amazon Mechanical Turk (AMT). These variants are characterized by the presence/absence of different explanation modalities used in train or test time. 
A demo of the interfaces is publicly available.\footnote{\url{http://deshraj.xyz/TOAIM/}}
\section{Vicki's Quirks}
\label{sec:vickis_quirks}
\begin{figure*}[ht!]
 \centering 
 \includegraphics[width=\textwidth,height=\textheight,keepaspectratio]{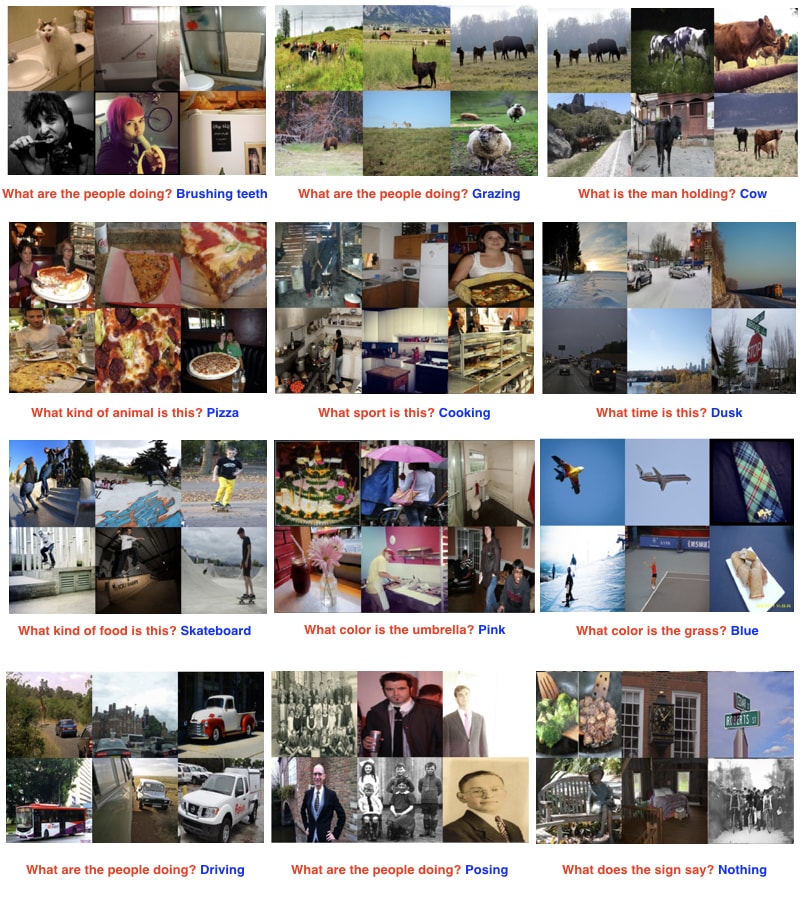}
 \caption{Given a question (red) we show images for which Vicki gave the same answer (blue) to the question to observe Vicki's quirks.  
 }
\label{fig:montages_1}
\end{figure*}

\begin{figure*}[ht!]
 \centering 
 \includegraphics[width=\textwidth]{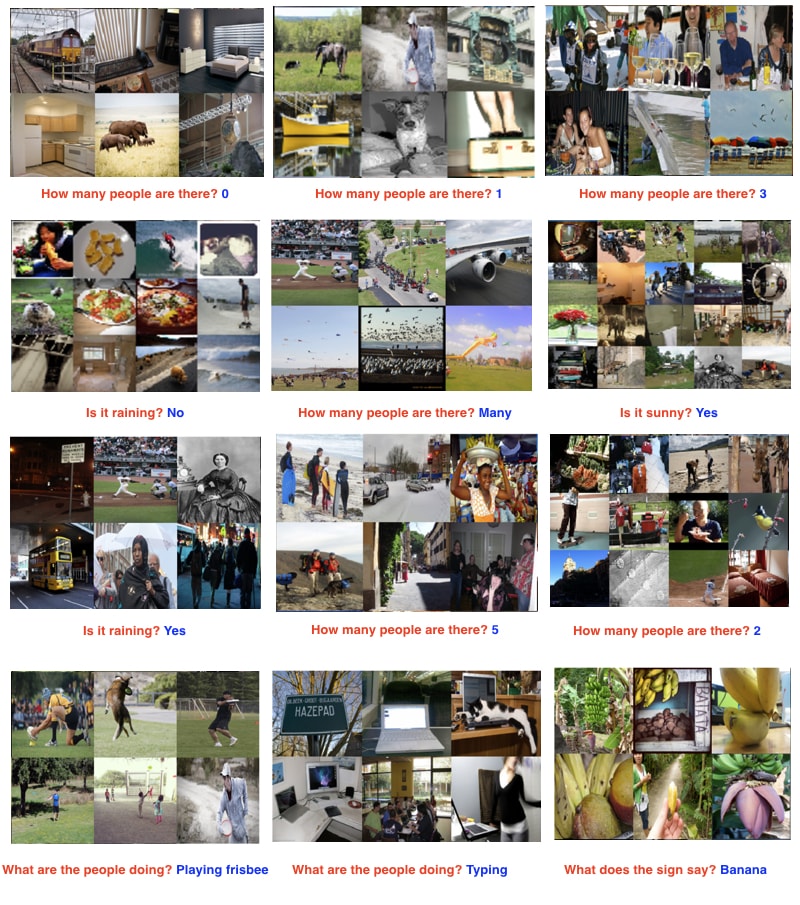}
 \caption{Given a question (red) we show images for which Vicki gave the same answer (blue) to the question to observe Vicki's quirks. 
 }
\label{fig:montages_2}
\end{figure*}

We present some examples in Fig.~\ref{fig:montages_1} and Fig.~\ref{fig:montages_2} that highlight Vicki's quirks. Recall that there are several factors which lead to Vicki being quirky, many of which are well known in VQA literature~\cite{agrawal2016analyzing}. As we can see across both examples, Vicki exhibits these quirks in a somewhat predictable fashion. At first glance, the primary factors that seem to decide Vicki's response to a question given an image are the properties and activities associated with the salient objects in the image, in combination with the language and the phrasing of the question being asked. This is evident when we look across the images (see Fig.~\ref{fig:montages_1} and~\ref{fig:montages_2}) for question-answer (QA) pairs such as -- \emph{What are the people doing? Grazing}, \emph{What is the man holding? Cow} and \emph{Is it raining? No}. As a specific example, notice the images for the QA pair \emph{What color is the grass? Blue} (see Fig.~\ref{fig:montages_1}) -- Vicki's response to this question is the most dominant color in the scene across all images even though there is no grass present in any of them. Similarly, for the QA pair \emph{What does the sign say? Banana} (see Fig.~\ref{fig:montages_2}) -- Vicki's answer is the salient object across all the scenes.

Interestingly, some subjects did try and pick up on some of the quirks and beliefs described previously, and formed a mental model of Vicki while completing the Failure Prediction or Knowledge Prediction tasks. We asked subjects to leave comments after completing a task and some of them shared their views on Vicki's behavior. We share some of those comments below. The abbreviations used are Failure Prediction (FP), Knowledge Prediction (KP) and Instant Feedback (IF).

\begin{figure}[ht!]
 \centering 
 \includegraphics[width=1\linewidth]{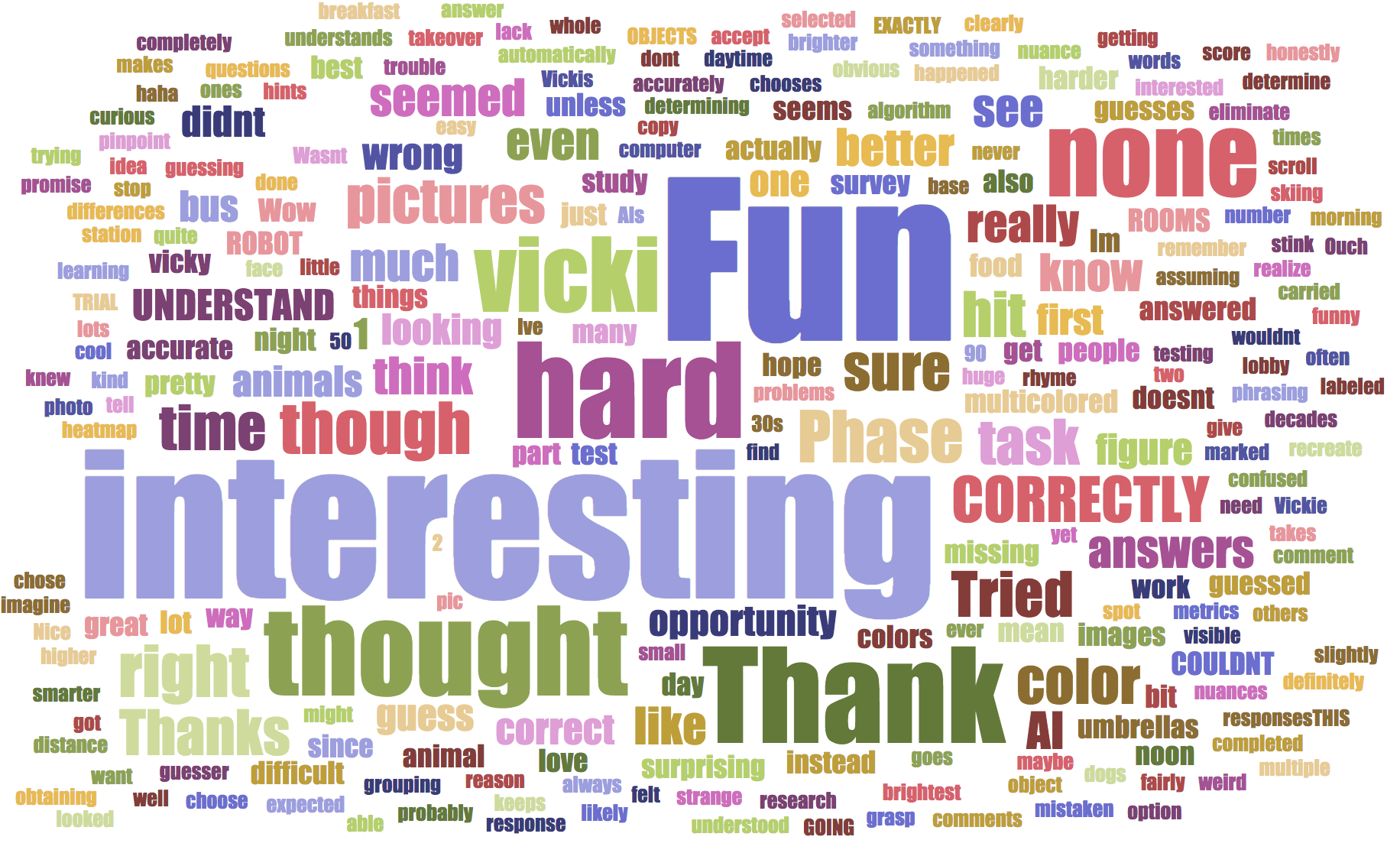}
 \caption{We show a word cloud of all the comments left by  subjects after completing the tasks across all settings. From the frequency of positive comments about the tasks, it appears that subjects were enthusiastic to familiarize themselves with Vicki.}
\label{fig:cmt_wcloud}
\end{figure}

\begin{compactenum}
\item \textbf{FP}
\begin{itemize}
\item \emph{These images were all pretty easy to see what animal it was. I would imagine the robot would be able to get 90\% of the animals correct, unless there were multiple animals in the same photo.}
\item \emph{I think the brighter the color the more likely they are to get it right. Multi-colored, not so sure.}
\item \emph{I'd love to know the answers to these myself.}
\end{itemize}
\item \textbf{FP + IF}
\begin{itemize}
\item \emph{This is fun, but kind of hard to tell what the hints mean. Can she determine the color differences in multi-colored umbrellas or are they automatically marked wrong because she only chooses one color instead of all of the colors? It seems to me that she just goes for the brightest color in the pic. This is very interesting. Thank you! :)}
\item \emph{I didn't quite grasp what the AI's algorithm was for determining right or wrong. I want to say that it was if the AI could see the face of the animal then it guessed correctly, but I'm really not sure.}
\end{itemize}
\item \textbf{FP + IF + Explanation Modalities}
\begin{itemize}
\item \emph{Even though Vicki is looking at the right spot doesn't always mean she will guess correctly. To me there was no rhyme or reason to guessing correctly. Thank you.}
\item \emph{I think she can accurately know a small number of people but cannot know a huge grouping yet.}
\item \emph{I would be more interested to find out how Vickis metrics work. What I was assuming is just color phase and distance might not be accurate.}
\end{itemize}
\item \textbf{KP}
\begin{itemize}
\item \emph{Time questions are tricky because all Vicki can do is round to the nearest number.}
\item \emph{there were a few that seemed like it was missing obvious answers - like bus and bus stop but not bus station.  Also words like lobby seemed to be missing.}
\end{itemize}
\item \textbf{KP + IF}
\begin{itemize}
\item \emph{Interesting, though it seems Vicki has a lot more learning to do. Thank you!}
\item \emph{This HIT was interesting, but a bit hard. Thank you for the opportunity to work this.}
\end{itemize}
\item \textbf{KP + IF + Explanation Modalities}
\begin{itemize}
\item \emph{You need to eliminate the nuances of night time and daytime from the computer and choose one phrasing "night" or "day" Vicki understands. The nuance keeps me and I'm sure others obtaining a higher score here on this task.}
\item \emph{I felt that Vickie was mistaken as to what some colors were for the first test which probably carried over and I tried my best to recreate her responses.}
\end{itemize}
\item \textbf{KP + IF + Montages}
\begin{itemize}
\item \emph{I am not sure that I ever completely understood how Vicki thought. It seemed it had more to do with what was in the pictures instead of the time of day it looked in the pictures. If there was food, she chose noon or morning, even though at times it was clearly breakfast food and she labeled it noon.}
\item \emph{It doesn't seem very accurate as I made sure to count and took my time assessing the pictures.}
\item \emph{it is hard to figure out what they are looking for since there isn't many umbrellas in the pictures}
\end{itemize}
\end{compactenum}

On a high-level reading through all comments, we found that subjects felt that Vicki's response often revolves around the most salient object in the image, that Vicki is bad at counting, and that Vicki often responds with the most dominant color in the image when asked a color question. In Fig.~\ref{fig:cmt_wcloud}, we show a word cloud of all the comments left by the subjects after completing the tasks. From the comments, we observed that subjects were very enthusiastic to familiarize themselves with Vicki, and found the process engaging. Many thought that the scenarios presented to them were interesting and fun, despite being hard. We used some basic elements of gamification, such as performance-based reward and narrative, to make our tasks more engaging; we think the positive response indicates the possibility of making such human-familiarization with AI engaging even in real-world settings. 
\section{Perception of AI}
\label{sec:public_perception}

In addition to measuring the subjects' capabilities to predict Vicki's behavior, we also conducted a survey to assess their general impressions of present-day AI. Specifically, we asked them to fill out a survey with questions focusing around three types of information - ``Background Information'', ``Familiarity with Computers and AI'' and ``Estimates of AI's capabilities''.

\begin{figure*}[t!]
\centering
\setlength{\fboxsep}{0pt}
\setlength{\fboxrule}{0pt}
\begin{subfigure}[t]{3in}
  	\fbox{\includegraphics[width=3in, height=2in]{supp_figures/comment_wordcloud.png}}
	\caption{We show a word cloud of all the comments left by  subjects after completing the tasks across all settings. From the frequency of positive comments about the tasks, it appears that subjects were enthusiastic to familiarize themselves with Vicki.}		
	\label{fig:cmt_wcloud}
\end{subfigure}	
\hspace{0.1in}    
\begin{subfigure}[t]{3in}
	\fbox{\includegraphics[width=3in, height=2in]{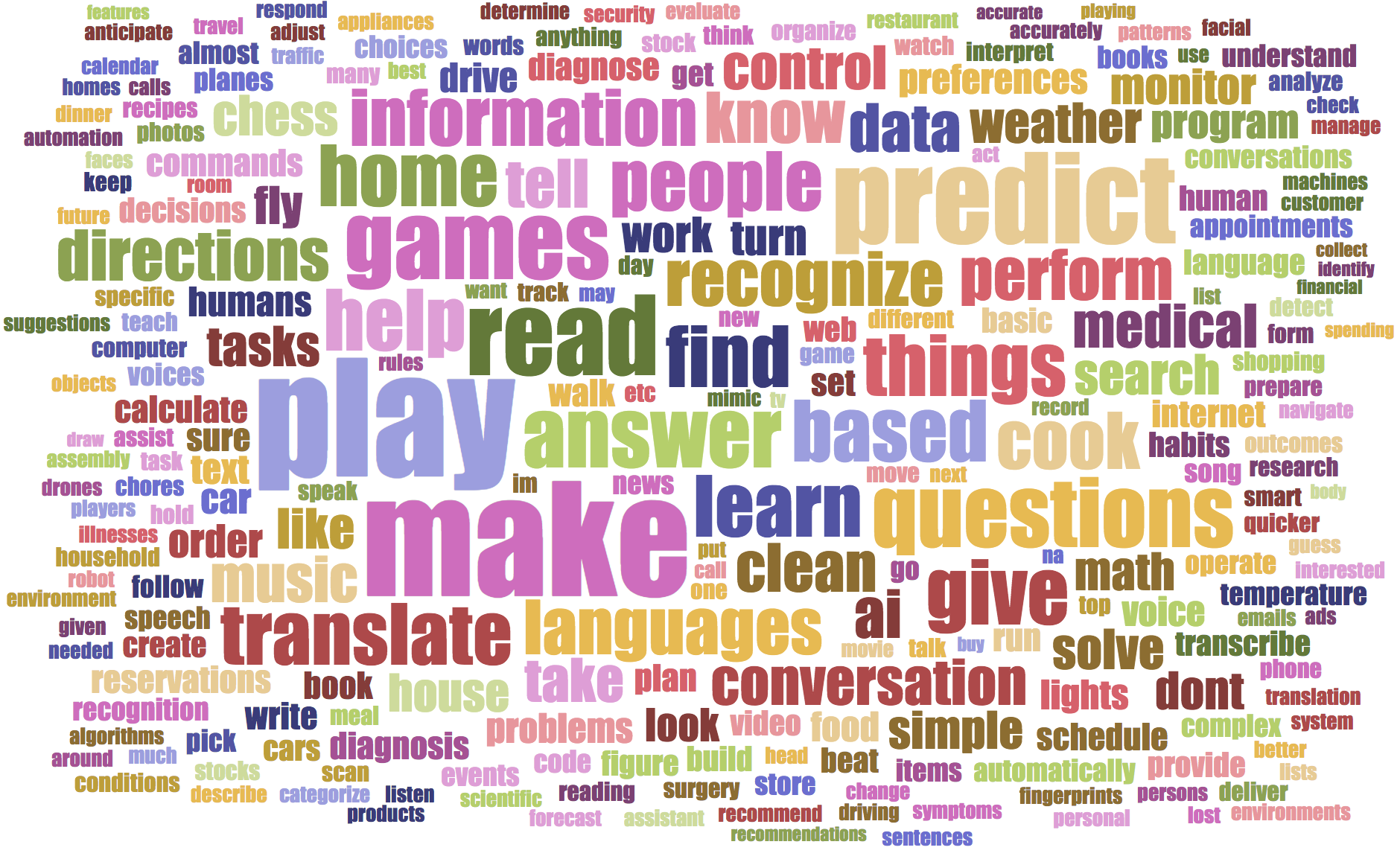}}
    \caption{A word cloud of subject responses to ``Name three things that you think AI today can do.''}
    \label{fig:ai_now}
\end{subfigure}
\begin{subfigure}[t]{3in}
	\fbox{\includegraphics[width=3in, height=2in]{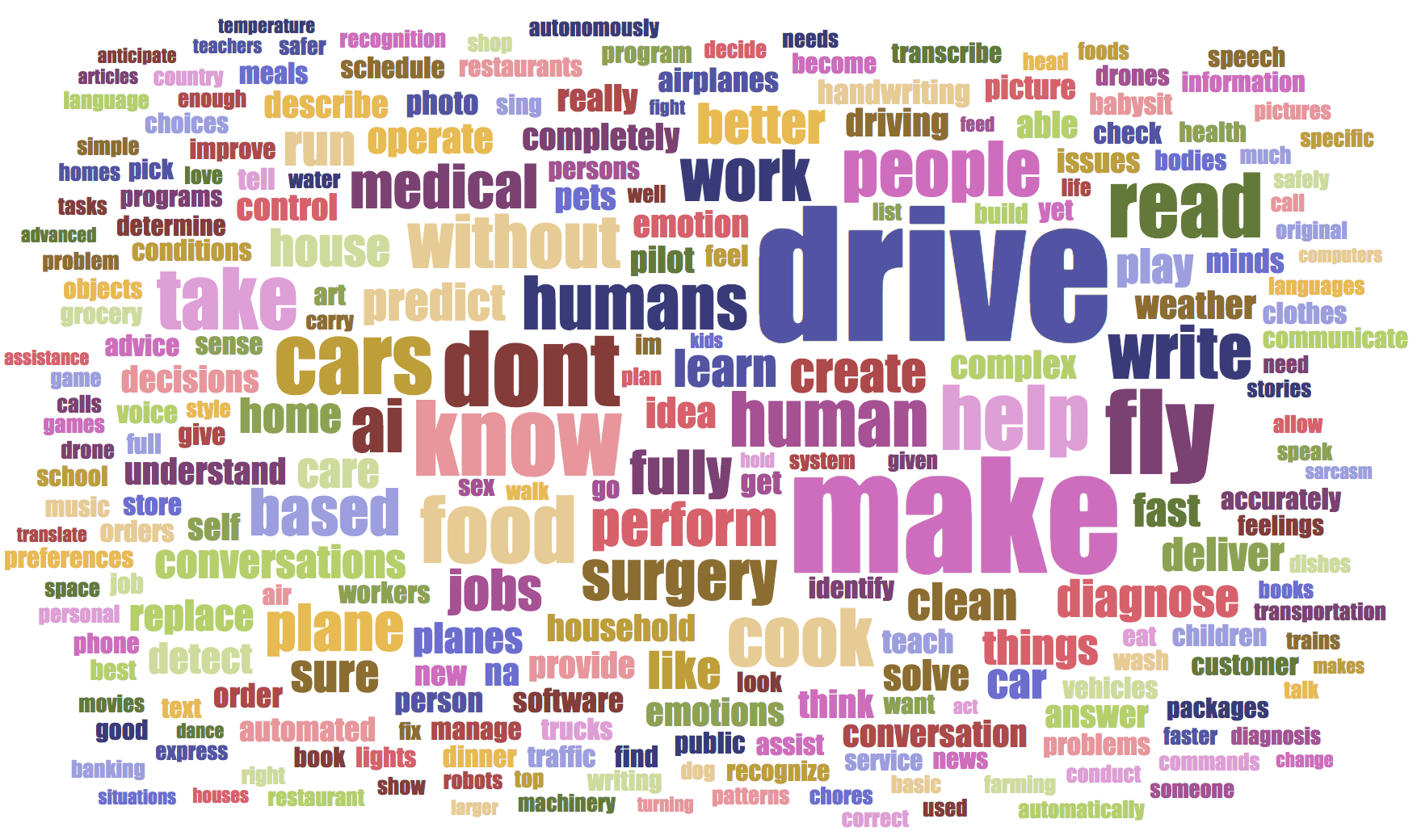}}
    \caption{A word cloud of subject responses to ``Name three things that you think AI today can't yet do but will be able to do in 3 years.''}
    \label{fig:ai_3_yrs}
\end{subfigure}         
\hspace{0.1in}    
\begin{subfigure}[t]{3in}
	\fbox{\includegraphics[width=3in, height=2in]{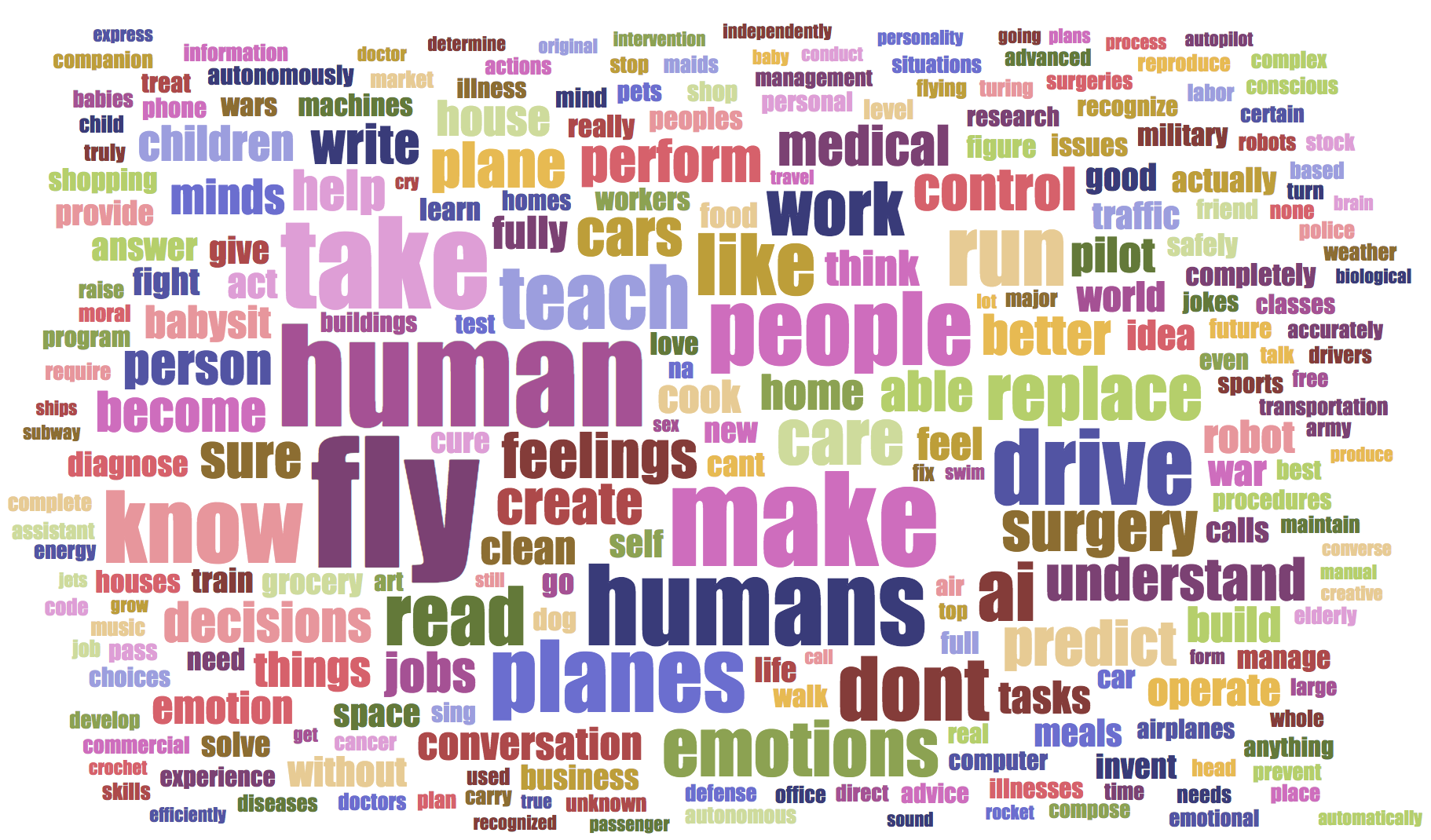}}
    \caption{A word cloud of subject responses to ``Name three things that you think AI today can't yet do and will take a while ($>$ 10 years) before it can do it.''}
	\label{fig:ai_10_yrs}
\end{subfigure}        
\begin{subfigure}[t]{3in}
	\fbox{\includegraphics[width=3in, height=2in]{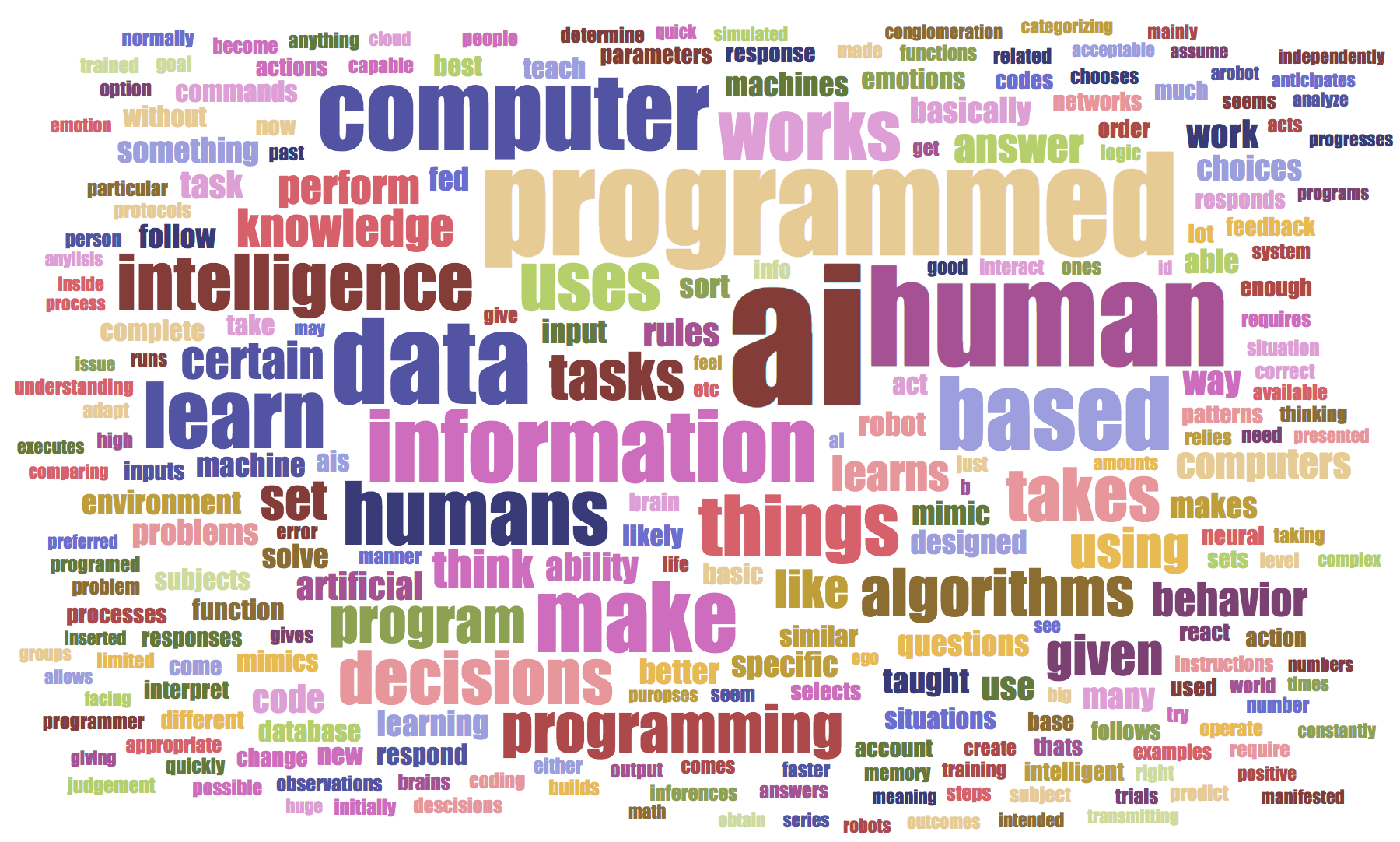}}
    \caption{A word cloud of subject responses when asked to describe how \emph{they} think AI today works. }
	\label{fig:how_ai_works}
\end{subfigure}
\caption{Word clouds corresponding to responses from humans for different questions.}
\label{fig:perception_of_ai}
\end{figure*}

\begin{figure*}[ht!]
 \centering 
 \includegraphics[width=1\textwidth]{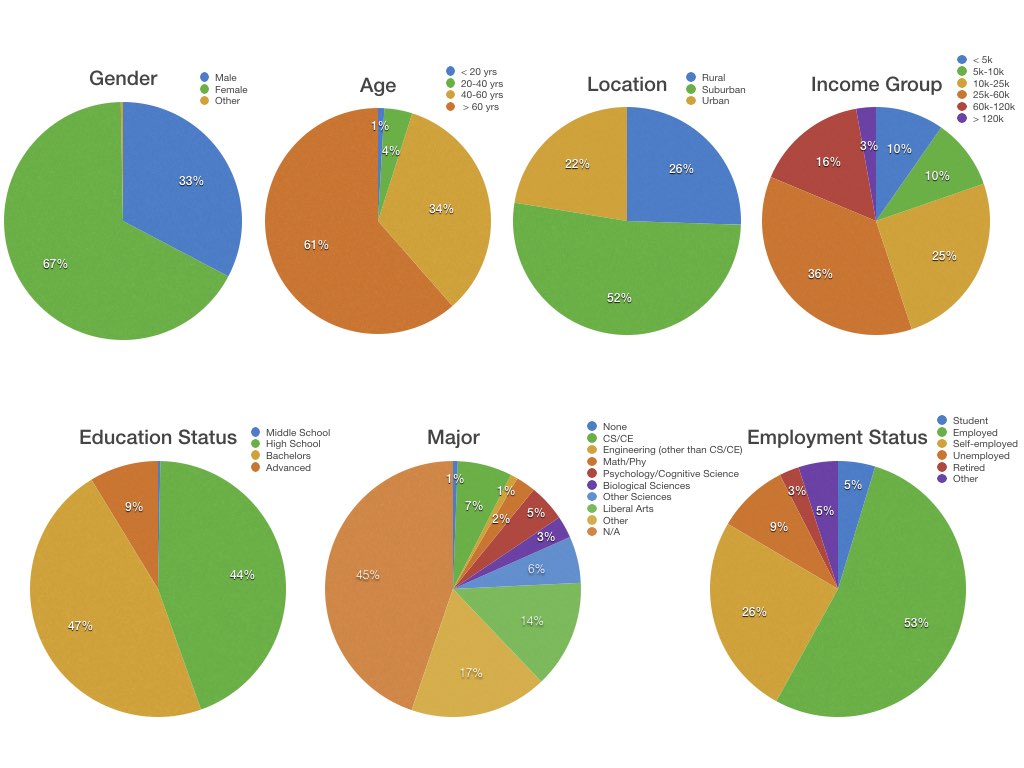}
 \vspace{-20pt}
 \caption{Population Demographics (across 321 subjects)}
 \vspace{-15pt}
\label{fig:pop_demog}
\end{figure*}

\begin{figure*}[ht!]
 \centering 
 \includegraphics[width=1\textwidth]{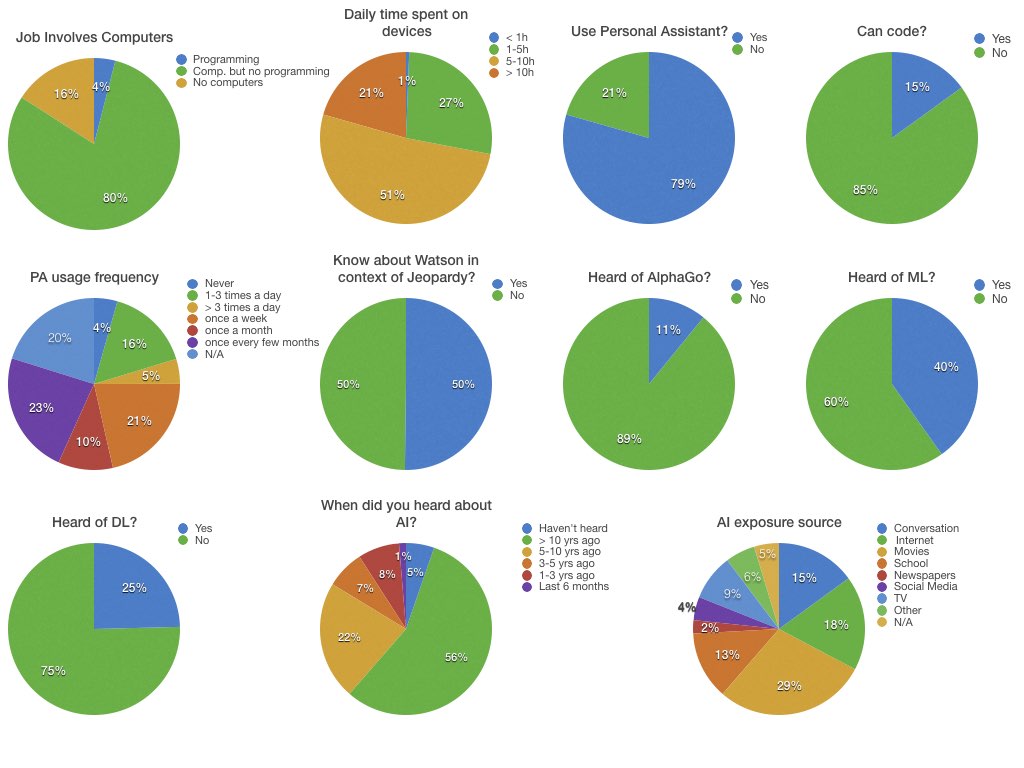}
 \vspace{-20pt}
 \caption{Technology and AI exposure (across 321 subjects)}	
 \vspace{-15pt}
\label{fig:tech_expo}
\end{figure*}

\begin{figure*}[ht!]
 \centering 
 \includegraphics[width=1\textwidth]{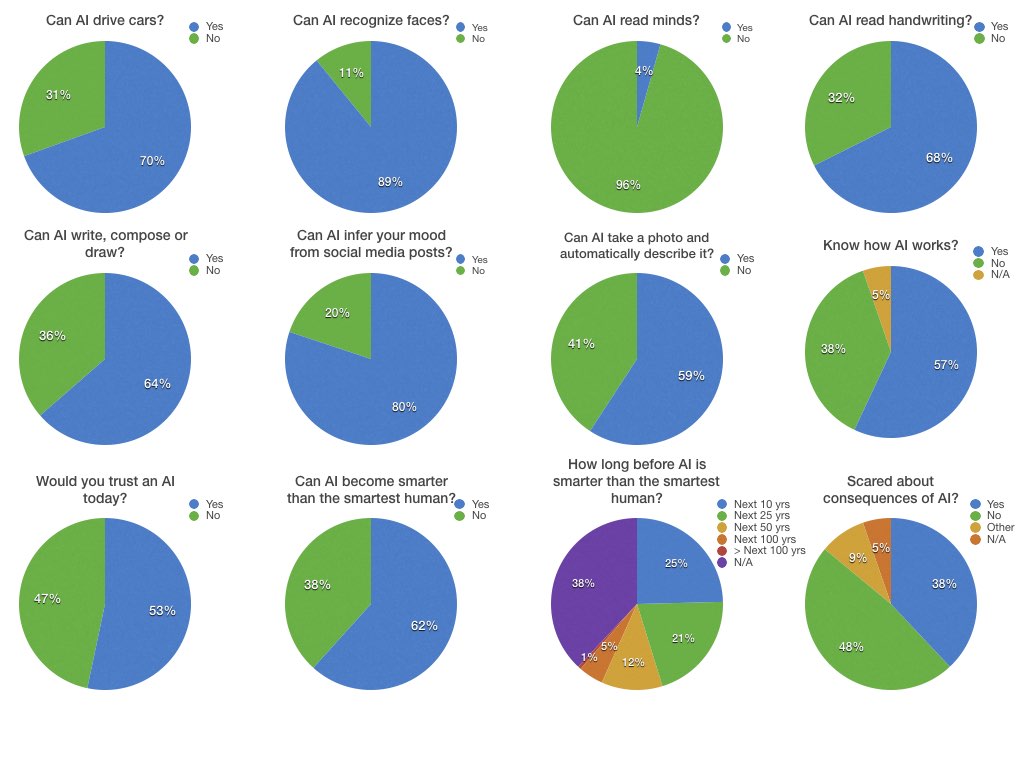}
 \vspace{-20pt}
 \caption{Perception of AI (across 321 subjects)}
 \vspace{-15pt}
\label{fig:ai_percep}
\end{figure*}

In Fig.~\ref{fig:pop_demog},~\ref{fig:tech_expo} and ~\ref{fig:ai_percep}, we break down the 321 subjects that completed the survey by their response to each question.

As part of the survey, subjects were also asked a few subjective questions about their opinions on present--day AI's capabilities. These include multiple-choice questions focusing on some specific capabilities of AI (``Can AI recognize faces?'', ``Can AI drive cars?'', etc.) -- responses to which are summarized in Fig.~\ref{fig:ai_percep}. 
The subjects were also asked to specifically list tasks that they thought AI is capable of performing \emph{today} (see Fig.~\ref{fig:ai_now}), will be capable of \emph{in the next 3 years} (see Fig.~\ref{fig:ai_3_yrs}), and will be capable of \emph{in the next 10 years} (see Fig.~\ref{fig:ai_10_yrs}). We also asked how \emph{they} think AI works (see Fig.~\ref{fig:how_ai_works}). In Fig.~\ref{fig:ai_now}, ~\ref{fig:ai_3_yrs} and ~\ref{fig:ai_10_yrs}, we show word clouds corresponding to what subjects thought about the capabilities of AI. We also share some of those responses below.

\begin{compactenum}
\item Name three things that you think AI today can do.
\emph{Predict sports games; Detect specific types of cancer in images; Control house temp based on outside weather; translate; calculate probabilities; Predictive Analysis; AI can predict future events that happen like potential car accidents; lip reading; code; Facial recognition; Drive cars; Play Go; predict the weather; Hold a conversation; Be a personal assistant; Speech recognition; search the web quicker.}

\item Name three things that you think AI today can't yet do but will be able to do in 3 years.
\emph{Fly planes; Judge emotion in voices; Predict what I want for dinner; perform surgery; drive cars; manage larger amounts of information at a faster rate; think independently totally; play baseball; drive semi trucks; Be a caregiver; anticipate  a person's lying ability; read minds; Diagnose patients; improve robots to walk straight; Run websites; solve complex problems like climate change issues; program other ai; guess ages; form conclusions based on evidence; act on more complex commands; create art.}

\item Name three things that you think AI today can't yet do and will take a while ($>$ 10 years) before it can do it.
\emph{Imitate humans; be indistinguishable from humans; read minds; Have emotions; Develop feelings; make robots act like humans; truly learn and think; Replace humans; impersonate people; teach; be a human; full AI with personalities; Run governments; be able to match a human entirely; take over the world; Pass a Turing test; be a human like friend; intimacy; Recognize things like sarcasm and humor.}
\end{compactenum}

Interestingly, we observe a steady progression in subjects' expectations of AI's capabilities, as the time span increases. On a high-level reading through the responses, we notice that subjects believe that AI today can successfully perform tasks such as \emph{machine translation}, \emph{driving vehicles}, \emph{speech recognition}, \emph{analyzing information and drawing conclusions}, etc. (see Fig.~\ref{fig:ai_now}). It is likely that this is influenced by the subjects' exposure to or interaction with some form of AI in their day-to-day lives. When asked about what AI can do three years from now, most subjects suggested more sophisticated tasks such as \emph{inferring emotions from voice tone}, \emph{performing surgery}, and even \emph{dealing with climate change issues} (see Fig.~\ref{fig:ai_3_yrs}). However, the most interesting trends emerge while observing subjects' expectation of what AI can achieve in the next 10 years (see Fig.~\ref{fig:ai_10_yrs}). A major proportion of subjects believe that AI will gain the ability to \emph{understand and emulate human beings}, \emph{teach human beings}, \emph{develop feelings and emotions} and \emph{pass the Turing test}. 

We also observe how subjects think AI works (see Fig.~\ref{fig:how_ai_works}). Mostly, subjects believe that an AI agent today is a system with high computational capabilities that has been programmed to simulate intelligence  and perform certain tasks by exposing it to huge amounts of information, or, as one of subjects phrased it -- \emph{broadly AI recognizes patterns and creates optimal actions based on those patterns towards some predefined goals}. In summary, it appears that subjects have high expectations from AI, given enough time. While it is uncertain at this stage how many, or how soon, these feats will actually be achieved, we believe that building a model of the AI's skillset will help humans generally become more active and effective collaborators in human--AI teams.


We now provide a full list of questions the subjects were asked in the survey.

\begin{compactenum}
\item How old are you?
	\begin{compactenum}
	\item Less than 20 years
    \item Between 20 and 40 years
    \item Between 40 and 60 years
    \item Greater than 60 years
	\end{compactenum}
\item What is your gender?
 	\begin{compactenum}
 	\item Male
    \item Female
    \item Other
 	\end{compactenum}
\item Where do you live?
 	\begin{compactenum}
 	\item Rural
 	\item Suburban
 	\item Urban
 	\end{compactenum}
\item Are you?
 	\begin{compactenum}
 	\item A student
 	\item Employed
 	\item Self-employed
 	\item Unemployed
 	\item Retired 
 	\item Other
 	\end{compactenum}
\item To which income group do you belong?
 	\begin{compactenum}
 	\item Less than 5000\$ per year
 	\item 5,000-10,000\$ per year
 	\item 10,000-25,000\$ per year
 	\item 25,000-60,000\$ per year 
 	\item 60,000-120,000\$ per year 
 	\item More than 120,000\$ per year
 	\end{compactenum}
\item What is your highest level of education?
 	\begin{compactenum}
 	\item No formal education
 	\item Middle School
 	\item High School
 	\item College (Bachelors)
 	\item Advanced Degree
 	\end{compactenum}
\item What was your major?
 	\begin{compactenum}
 	\item Computer Science / Computer Engineering
 	\item Engineering but not Computer Science
 	\item Mathematics / Physics
 	\item Philosophy
 	\item Biology / Physiology / Neurosciences
 	\item Psychology / Cognitive Sciences
 	\item Other Sciences
 	\item Liberal Arts
 	\item Other
 	\item None
 	\end{compactenum}
\item Do you know how to program / code?
 	\begin{compactenum}
 	\item Yes
 	\item No
 	\end{compactenum}
\item Does your full-time job involve:
 	\begin{compactenum}
 	\item No computers
 	\item Working with computers but no programming / coding?
 	\item Programming / Coding
 	\end{compactenum}
\item How many hours a day do you spend on your computer / laptop / smartphone?
 	\begin{compactenum}
 	\item Less than 1 hour
 	\item 1-5 hours
 	\item 5-10 hours
 	\item Above 10 hours
 	\end{compactenum} 
\item Do you know what Watson is in the context of Jeopardy?
 	\begin{compactenum}
 	\item Yes
 	\item No
 	\end{compactenum}
\item Have you ever used Siri, Alexa, or Google Now/Google Assistant?
 	\begin{compactenum}
 	\item Yes
 	\item No
 	\end{compactenum}
\item How often do you use Siri, Alexa, Google Now, Google Assistant, or something equivalent?
 	\begin{compactenum}
 	\item About once every few months
 	\item About once a month
 	\item About once a week
 	\item About 1-3 times a day
 	\item More than 3 times a day
 	\end{compactenum}
\item Have you heard of AlphaGo?
 	\begin{compactenum}
 	\item Yes
 	\item No
 	\end{compactenum}
\item Have you heard of Machine Learning?
 	\begin{compactenum}
 	\item Yes
 	\item No
 	\end{compactenum}
\item Have you heard of Deep Learning?
 	\begin{compactenum}
 	\item Yes
 	\item No
 	\end{compactenum}
\item When did you first hear of Artificial Intelligence (AI)?
	\begin{compactenum}
	\item I have not heard of AI
	\item More than 10 years ago
	\item 5-10 years ago
	\item 3-5 years ago
	\item 1-3 years ago
	\item In the last six months
	\item Last month
	\end{compactenum}
\item How did you learn about AI?
	\begin{compactenum}
	\item School / College
	\item Conversation with people
	\item Movies
	\item Newspapers
	\item Social media
	\item Internet
	\item TV
	\item Other
	\end{compactenum}
\item Do you think AI today can drive cars fully autonomously?
	\begin{compactenum}
	\item Yes
	\item No
	\end{compactenum}
\item Do you think AI today can automatically recognize faces in a photo?
	\begin{compactenum}
	\item Yes
	\item No
	\end{compactenum}
\item Do you think AI today can read your mind?
	\begin{compactenum}
	\item Yes 
	\item No
	\end{compactenum}
\item Do you think AI today can automatically read your handwriting?
	\begin{compactenum}
	\item Yes
	\item No
	\end{compactenum}
\item Do you think AI today can write poems, compose music, make paintings?
	\begin{compactenum}
	\item Yes
	\item No
	\end{compactenum}
\item Do you think AI today can read your Tweets, Facebook posts, etc. and figure out if you are having a good day or not?
	\begin{compactenum}
	\item Yes
	\item No
	\end{compactenum}
\item Do you think AI today can take a photo and automatically describe it in a sentence?
	\begin{compactenum}
	\item Yes
	\item No
	\end{compactenum}
\item Other than those mentioned above, name three things that you think AI today can do.
\item Other than those mentioned above, name three things that you think AI today can't yet do but will be able to do in 3 years.
\item Other than those mentioned above, name three things that you think AI today can't yet do and will take a while ($>$ 10 years) before it can do it.
\item Do you have a sense of how AI works?
	\begin{compactenum}
	\item Yes
	\item No
	\item If yes, describe in a sentence or two how AI works.
	\end{compactenum}
\item Would you trust an AI's decisions today?
	\begin{compactenum}
	\item Yes
	\item No
	\end{compactenum}
\item Do you think AI can ever become smarter than the smartest human?
	\begin{compactenum}
	\item Yes
	\item No
	\end{compactenum}
\item If yes, in how many years?
	\begin{compactenum}
	\item Within the next 10 years
	\item Within the next 25 years
	\item Within the next 50 years
	\item Within the next 100 years
	\item In more than 100 years
	\end{compactenum}
\item Are you scared about the consequences of AI?
	\begin{compactenum}
	\item Yes
	\item No
	\item Other
	\item If other, explain.
	\end{compactenum}
\end{compactenum}
\vspace{-5pt}

\end{document}